\documentclass{article}

\usepackage[nonatbib, preprint]{neurips_2023}

\usepackage[utf8]{inputenc} % allow utf-8 input
\usepackage[T1]{fontenc}    % use 8-bit T1 fonts
\usepackage{hyperref}       % hyperlinks
\usepackage{url}            % simple URL typesetting
\usepackage{booktabs}       % professional-quality tables
\usepackage{amsfonts}       % blackboard math symbols
\usepackage{nicefrac}       % compact symbols for 1/2, etc.
\usepackage{microtype}      % microtypography
\usepackage{xcolor}         % colors

\usepackage{graphicx}
\usepackage{amsmath}
\usepackage[font=small, labelfont=small]{caption}
\usepackage{subcaption}
\usepackage{overpic}
\usepackage{physics}
\usepackage{color}
\usepackage{soul}
\usepackage{wrapfig}
\usepackage[numbers,sort&compress]{natbib}
\usepackage{cleveref}
\usepackage{booktabs}
\usepackage{array}
\usepackage{multirow}
\usepackage{overpic}
\usepackage[final]{pdfpages}

\newcommand{\PreserveBackslash}[1]{\let\temp=\\#1\let\\=\temp}
\newcolumntype{C}[1]{>{\PreserveBackslash\centering}p{#1}}
\newcolumntype{R}[1]{>{\PreserveBackslash\raggedleft}p{#1}}
\newcolumntype{L}[1]{>{\PreserveBackslash\raggedright}p{#1}}

\definecolor{darkgreen}{rgb}{0.0,0.5,0}
\definecolor{darkblue}{rgb}{0.1,0.2,0.8}

\newcommand{\R}{\mathbb{R}}
\newcommand{\E}{\mathbb{E}}
\newcommand{\loss}[1]{\mathcal{L}_{\textrm{#1}}}

\usepackage{paralist}  % tighter enumeration
\setdefaultleftmargin{2.5em}{2em}{1.7em}{1.5em}{1em}{1em}

\newcommand{\geonet}{{\mathcal{G}}}
\newcommand{\texture}{{\mathcal{T}}}
\newcommand{\decoder}{{\mathcal{D}}}

\newcommand{\prompt}{{p}}

\newcommand{\rasterize}{{\mathcal{R}}}
\newcommand{\azimuth}{{a}}
\newcommand{\elevation}{{e}}
\DeclareMathOperator{\lap}{L}
\DeclareMathOperator{\lbs}{LBS}
\newcommand{\shape}{\beta}
\newcommand{\expression}{\psi}
\newcommand{\pose}{\phi}
\newcommand{\flame}{\mathcal{F}}
\newcommand{\blend}{\mathcal{B}}
\newcommand{\featureimage}{F}
\newcommand{\maskimage}{M}
\newcommand{\image}{I}

\usepackage{xpunctuate}
\providecommand{\eg}[0]{e.g\xperiod}
\providecommand{\ie}[0]{i.e\xperiod}

\providecommand{\wrt}[0]{w.r.t\xperiod}

\crefname{section}{Sec.}{Secs.}
\Crefname{section}{Section}{Sections}
\Crefname{table}{Table}{Tables}
\crefname{table}{Tab.}{Tabs.}
\Crefname{figure}{Figure}{Figures}
\crefname{figure}{Fig.}{Figs.}
\creflabelformat{equation}{#2#1#3}

\newcommand{\ours}{our method}

\title{Articulated 3D Head Avatar Generation using Text-to-Image Diffusion Models}

\author{%
  Alexander W. Bergman \\
  Stanford University\\
	\texttt{awb@stanford.edu}\\
  \And
  Wang Yifan\\
  Stanford University\\
	\texttt{yifan.wang@stanford.edu}\\
  \And
  Gordon Wetzstein\\
  Stanford University\\
	\texttt{gordonwz@stanford.edu}\\
}

\begin{document}

\maketitle
\vbox{%
	\vskip -0.15in
	\hsize\textwidth
	\linewidth\hsize
	\centering
	\normalsize
	%	\vskip 0.1in
	\tt\href{https://www.computationalimaging.org/publications/articulated-diffusion/}{computationalimaging.org/publications/articulated-diffusion/}
	\vskip 0.28in
}

\begin{abstract}
The ability to generate diverse 3D articulated head avatars is vital to a plethora of applications, including augmented reality, cinematography, and education. Recent work on text-guided 3D object generation has shown great promise in addressing these needs. These methods directly leverage pre-trained 2D text-to-image diffusion models to generate 3D-multi-view-consistent radiance fields of generic objects.
However, due to the lack of geometry and texture priors, these methods have limited control over the generated 3D objects, making it difficult to operate inside a specific domain, \eg, human heads.
In this work, we develop a new approach to text-guided 3D head avatar generation to address this limitation.
Our framework directly operates on the geometry and texture of an articulable 3D morphable model (3DMM) of a head, and introduces novel optimization procedures to update the geometry and texture while keeping the 2D and 3D facial features aligned.
The result is a 3D head avatar that is consistent with the text description and can be readily articulated using the deformation model of the 3DMM.
We show that our diffusion-based articulated head avatars outperform state-of-the-art approaches for this task. The latter are typically based on CLIP, which is known to provide limited diversity of generation and accuracy for 3D object generation.
\end{abstract}

\section{Introduction}
\label{sec:intro}
The generation of high-quality, editable 3D head avatars is a long-standing research problem spanning computer graphics and machine learning.
Recent advances in generative 3D networks have been successful in learning to generate close-to-photorealistic digital humans from unstructured collections of single-view images (see \cref{sec:related_work}).
The diversity of assets generated with these methods, however, is limited to individual object classes and the limited amount of training data.
Instead, we seek to unlock the immense potential of pre-trained 2D text-to-image foundation models to directly help generate 3D articulable head avatars, greatly expanding both the control over and span of possible generated results.

Recent text-guided 3D object generation approaches leverage pre-trained 2D diffusion models to generate 3D assets~\cite{poole2022dreamfusion,wang2022score,lin2022magic3d,metzer2022latent}.
These methods use radiance fields as the underlying representation, which are not tied to a head prior. As such, these methods require additional steps to be constrained for 3D avatar generation and the generated avatars are also not easy to articulate due to the intrinsic limitation of the representation.
Text-guided 3D head avatar generation addresses this limitation by generating neural textures on parametric head template shapes~\cite{michel2022text2mesh,canfes2023text,aneja2022clipface}. These methods successfully build on the 2D embedding space of contrastive language--image pre-training (CLIP)~\cite{radford2021learning}, though 3D objects produced by this approach tend to lack the diversity and accuracy with respect to text prompts~\cite{jain2022zero,poole2022dreamfusion}.

In this work, we propose to use the prior encoded in state-of-the-art 2D text-to-image diffusion models as a proxy source of data for learning to generate representations of 3D head avatars.
For this task, we leverage the score distillation loss~\cite{poole2022dreamfusion, wang2022score} to learn a 3D-view-consistent representation guided by an input text prompt.
Given a parametric head model~\cite{FLAME:SiggraphAsia2017} at initialization, our training scheme optimizes both shape and appearance of the avatar from the prompt. Specifically, shape is optimized by refining the vertex positions of the template mesh and appearance is modeled as a texture map of feature embeddings that are decoded into the differentiably rendered images using a latent diffusion model decoder~\cite{rombach2021highresolution}. Avatars generated with our method can be easily animated by articulating the parametric template.

To our knowledge, this is the first approach to leverage emerging text-to-2D-image diffusion models to jointly optimize the shape and appearance to generate an articulable 3D head avatar.
Our framework shows promise in generating a diversity of 3D heads which are difficult to synthesize using existing generative models, including fictional humanoids.
Our approach is a crucial step towards accelerating the 3D content creation process. This is crucial in applications such as gaming or cinematography, where artists undergo a laborious process to manually generate and rig meshes of characters.

To summarize, the contributions of our approach are:
\begin{compactitem}
	\item We present a method for generating 3D-view-consistent and articulable human head avatars from only a text prompt using state-of-the-art text-to-2D-image diffusion models. To the best of our knowledge, this is the first approach of its kind.
	\item We develop a representation and optimization schedule which allows for both shape and appearance of an explicit representation to be optimized to be consistent with a text prompt.
	\item We demonstrate high-quality head avatar generation results from a variety of text prompts, outperforming previous baselines in generating diverse but semantically meaningful results which can easily be articulated.
\end{compactitem}

\begin{figure}[t!]
	\includegraphics[width=\textwidth]{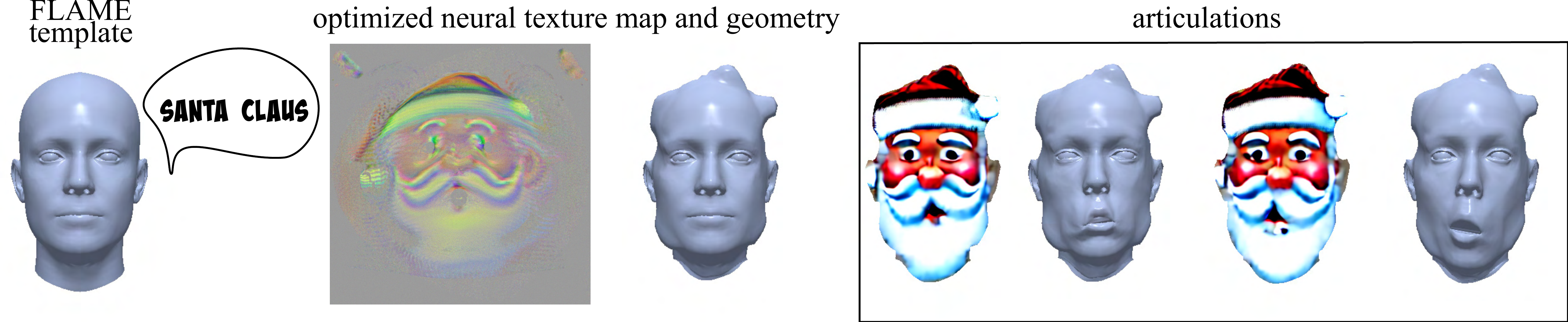}
	\caption{We propose a text-based 3D head avatar generation method using 2D diffusion models. From a parametric head model~\cite{FLAME:SiggraphAsia2017}, we optimize its geometry and texture to fit a text prompt. Thanks to the parametric nature of the template model, the generated head avatar can be easily articulated.}
	\label{fig:teaser}
\end{figure}

\section{Related Work}
\label{sec:related_work}
\paragraph{Modeling, Generating, and Animating Human Heads.}

Parametric 3D morphable face models (3DMMs) have been a staple of the computer graphics and vision communities for the past two decades~\cite{egger20203d}.
Advances in neural rendering, often combined with 3DMMs, have recently enabled articulated 3D shape and appearance fitting of a human head from one or multiple images~\cite{tewari2021state, grassal2022neural}.
3D-aware generative adversarial networks (GANs)  represent another suite of techniques that can learn 3D representations of digital humans from unstructured collections of single-view images~\cite{Szabo:2019,hologan,graf,pigan,cips3d,eg3d,volumegan,skorokhodov2022epigraf,stylesdf,gram,xiang2022gram,zhao2022generative,shadegan,sun2021fenerf,d3d,zhang2022multi,chen2022sofgan,zhang20223d,stylenerf,giraffe,xue2022giraffe,schwarz2022voxgraf,an2023panohead,deng2023lumigan}. Recent variants of these approaches also utilize 3DMMs to allow the generated assets to be freely articulated~\cite{grigorev2021stylepeople,bergman2022gnarf,noguchi2022unsupervised,yang20223dhumangan,sun2023next3d,jiang2023humangen,Avatargen2023,hong2023evad}. None of these methods, however, allow the text-guided head generation we aim to enable and many of them are limited to very specific training data and do not generalize beyond this.

\paragraph{Text-guided 3D Generation.}

Text-guided 3D generation approaches typically build on CLIP~\cite{radford2021learning} (e.g.,~\cite{jain2022zero,wang2021clip,khalid2022clipmesh}) or 2D diffusion methods~\cite{song2019generative,ho2020denoising,balaji2022ediffi,rombach2022high,saharia2022photorealistic} for 3D settings~\cite{poole2022dreamfusion,wang2022score,lin2022magic3d,metzer2022latent}.
All of these methods use some form of radiance field as a representation, which is designed for class-agnostic object generation, and thus lack articulable control.
While this issue can be mitigated by a additional step to align the density field to a human parametric mesh template, as proposed in~\cite{hong2022avatarclip,cao2023dreamavatar, jiang2023avatarcraft}, radiance fields are known for costly rendering and deformation, making them a less attractive choice in practice.
Closest to our work are the seminal works of Text2Mesh~\cite{michel2022text2mesh}, Latent3D~\cite{canfes2023text}, and ClipFace~\cite{aneja2022clipface}. All of these methods generate a texture, and some also vertex offsets; the latter two further utilize a template 3DMM of human head to specialize in 3D head avatar generation.
We consider ClipFace the state-of-the-art (SOTA) approach to text-guided 3D head generation as it demonstrated to outperform other relevant baselines.
All of these baselines, however are based on CLIP, which has been shown to offer limited variety and realism when used for 3D object generation~\cite{jain2022zero,poole2022dreamfusion}.

Also related to our approach are the concurrently and independently developed methods DreamFace~\cite{zhang2023dreamface} and Zero-Shot Text-to-Parameter Translation~\cite{zhao2023zeroshot}. DreamFace uses a 3DMM and texture to generate 3D faces with text.
Our method differs from theirs in that we do not use large high-quality asset collection for training. Zero-Shot Text-to-Parameter Translation uses a parametric representation in a game engine, while our method is able to generate results without access to this specific prior.
Additionally, the concurrent works DreamAvatar~\cite{cao2023dreamavatar} and AvatarCraft~\cite{jiang2023avatarcraft} also focus on the creation of articulable avatars from text prompts.  
However, these works both focus on human bodies, and thus support body articulation rather than facial expression manipulation. Moreover, DreamAvatar~\cite{cao2023dreamavatar} requires training a new representation for each desired body pose, while our method optimizes a single representation which can be controlled by a 3DMM.

\section{Text to 3D Head Avatar}
\label{sec:method}
\begin{figure}[t]
    \centering
\includegraphics[width=\linewidth]{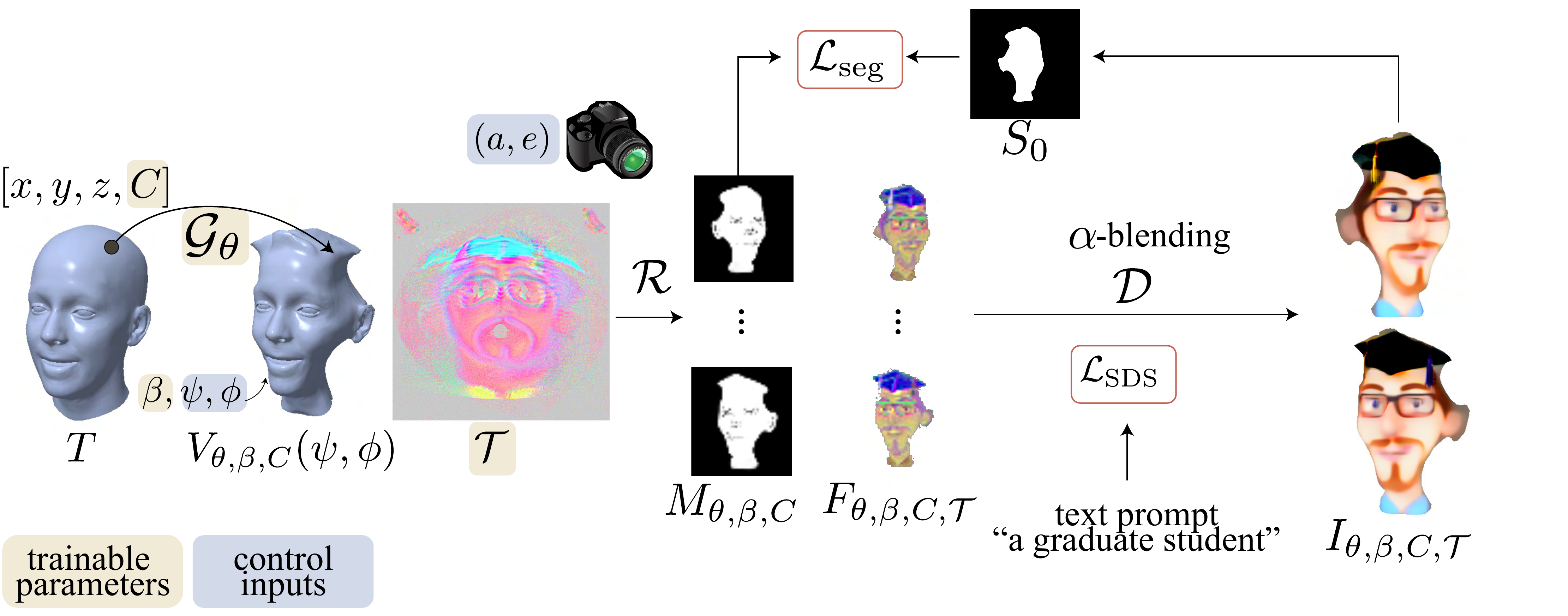}
    \caption{\textbf{Overview.} We represent a 3D avatar using a textured 3DMM (\cref{Sec:Representation}). Given a text prompt, we optimize for a neural texture map $\texture$, as well as the geometry through a geometry network \(\geonet_{\theta}\), the shape blendshape parameters \(\shape\) of the 3DMM, and per-vertex features \(C\) (see \cref{Sec:SDS}). During each optimization step, under sampled expression \(\expression\), joint locations \(\pose\), and camera angles \(\left( \azimuth, \elevation \right)\), we use a differentiable mesh rasterizer \(\rasterize\) to produce a feature map \(\featureimage\), which is then augmented with our \(\alpha-\)blending described in \cref{sec:training} before being decoded through a latent diffusion decoder \(\decoder\) to the RGB image \(\image\). Using Score Distillation Sampling (SDS), we obtain gradients from the diffusion network to match the trainable scene parameters to the given text prompt. Additionally, we propose a novel segmentation loss to ensure texture-geometry alignment (see \cref{sec:training}).}
    \label{fig:Overview}
\end{figure}
First, we describe the representation and rendering of our articulated 3D heads avatars in \cref{Sec:Representation}.
Then we briefly recap the score distillation approach for optimizing our model based on text prompts in \cref{Sec:SDS}.
Finally, in \cref{sec:training}, we propose a set of critical techniques to improve the geometry-consistency of the synthesized textures, leading to more realistic articulation.
For additional implementation details, please see our supplementary document.

\subsection{Articulated 3D head representation.}
\label{Sec:Representation}

We adopt a textured mesh with per-vertex offsets as our articulable head avatar representation.
Compared to other 3D representations based on radiance fields~\cite{mildenhall2020nerf}, our representation is automatically rigged to a parametric template head model, and does not require learning a continuous 3D deformation function in order to model pose-conditioned deformations.

\paragraph{Shape.}
We use a human head 3DMM, FLAME~\cite{FLAME:SiggraphAsia2017}, to represent the geometry.
Starting with a template mesh \(T\), FLAME first applies blendshapes using shape $\shape$, expression \(\expression\), and pose parameters \(\pose\), followed by a deformation to a new pose using standard linear blend skinning (LBS).
This model can be written as: \(\flame\left( \shape,\expression,\pose \right) = \lbs\left( T + \blend\left(\shape, \expression, \pose \right) \right) \),
where \(B\) denotes the blending of expression and pose, and \(\lbs\) denotes the LBS deformation.

To address the issue that FLAME does not capture hair and other possible head accessories, we use an MLP \(\geonet_{\theta}:\R^{3+d}\rightarrow\R^{3}\) to model per-vertex offsets from the flame template \(T\).
Specifically, \(\geonet_{\theta}\) takes the concatenation of the vertex positions of the initial FLAME template \(T\) and \(d-\)dimensional per-vertex features \(C\) as input, and outputs the per-vertex offsets.
Thus, vertex positions for a given pose in our representation are defined as:
\begin{equation}
    V_{\theta,\shape,C}(\expression,\pose) = \lbs\left( \blend\left( \shape, \expression, \pose \right) + \geonet_\theta(\left\lbrack T; C\right\rbrack)\right)\label{eq:shape_offset}.
\end{equation}

For a given text prompt's geometry, we optimize $\theta$, $\shape$, and $C$ creating a modified geometry FLAME template which can be controlled by $\expression$ and $\pose$.

While it would be possible to directly optimize these offsets, we find that using an MLP to represent offsets as a function of vertex position and features leads to more stable optimization due to the spectral bias of MLPs~\cite{tancik2020fourier, sitzmann2020siren} towards smooth functions.

\paragraph{Appearance.}
We model the appearance of head avatars using a 4-channel neural texture map $\texture$ of resolution \(512\times512\)~\cite{rombach2021highresolution}.
A differentiable mesh rasterizer \(\rasterize\) based on SoftRasterizer~\cite{liu2019soft} is used to render the textured mesh at a given camera pose into a 4-channel feature image $\featureimage$ of resolution \(64\times64\), along with a soft silhouette mask $\maskimage$ of the same resolution.

Finally the rendered feature image is decoded to the final RGB image of \(512\times 512\) using the latent decoder, \(\decoder\) from the pre-trained diffusion model.
Formally, the final image formation function is defined below, where the entities in the subscript are optimization variables, and the entities in the parentheses are given during training:
\begin{align}
\image_{\theta, \shape, C, \texture}\left( \pose,\expression, \azimuth, \elevation \right) &= \decoder\left( \featureimage \right),\\
\featureimage_{\theta, \shape, C, \texture}\left( \pose,\expression, \azimuth, \elevation \right) &= \rasterize\left( V_{\theta,\shape,C}\left(\expression,\pose\right), \texture, \azimuth, \elevation \right).
\end{align}

Both the vertex offsets and texture map are shown in ~\cref{fig:Overview}, along with the differentiable rendering process.

\subsection{Diffusion-guided representation optimization}
\label{Sec:SDS}

Given a text prompt \(\prompt\), our goal is to optimize the scene parameters such that the head avatar rendered with using arbitrary expression, pose, and camera positions are consistent with the text prompt.
Several prior works~\cite{poole2022dreamfusion, wang2022score, zhou2022sparsefusion} have studied this problem for generic 3D scene generation or reconstruction, arriving at similar loss construct from different derivations.
In this work, we adopt Score Distillation Sampling (SDS) and recap the formulation briefly below.

The SDS converts a conditional diffusion model into an image--text consistency loss.
Specifically, the diffusion model computes a gradient of this loss, which is then propagated back through the feature image $\featureimage$ to scene parameters used to generate it, namely \(\theta, \shape, C, \texture\).
This gradient is computed as
\begin{equation}
    \nabla_{\featureimage}\loss{SDS}(\featureimage) = \mathbb{E}_{t,\epsilon} \left[ w^{-1}\left( \Gamma\left( \featureimage+w\epsilon,\prompt \right) - \epsilon \right)\right],\label{eq:score_jacobian_chaining}
\end{equation}
where \(w\) and \(\epsilon\sim\mathcal{N}\left( \mathbf{0}, \mathbf{I} \right)\) are the diffusion step size and noise respectively, and \(\Gamma\) is the diffusion model architecture which predicts the noise added to the latent.
%\yifan{Alex, please check if \cref{eq:score_jacobian_chaining} is correct.}

During optimization, we randomly sample the control inputs \(\expression, \pose, \azimuth, \) and \(\elevation\),
encouraging the scene parameters to produce an identity-consistent avatar that match the text prompt under varying expression, poses and camera angles.

\subsection{Geometry-consistent dual optimization.}\label{sec:training}
While using the above optimization objective with SDS can produce reasonable renderings, the lack of geometric constraints leads to misaligned geometry and texture (see \cref{sec:ablation}).
This is because the SDS loss operates only in image space, leaving the model to change the texture freely without changing the updating underlying geometry.
Such misalignment can cause uncanny articulation artifacts with the predefined blendshapes and skinning weights, nullifying the benefits of using a 3DMM.
To combat this issue, we propose an optimization procedure between texture and geometry as described below, which significantly improves the geometry and texture alignment, hence producing more convicing animations.

\paragraph{Geometry-aware texture-only optimization.} We begin with a texture-only optimization step on an enlarged FLAME template, while fixing \(\geonet_{\theta}\) and \(\beta\), and \(C\).
Enlarging the FLAME template simply provides more canvas for the texture to paint on. We implement this step by applying Laplacian Smoothing~\cite{sorkine2004laplacian} with a negative strength.
In order to encourage the painted texture to match with facial features, we render the mesh \( V_{\theta, \shape, C}\left( \expression, \pose \right) \) \emph{without texture} under a fixed directional lighting and linearly blend this textureless image with the rendered feature image at each step using a blending parameter $\alpha$.
We empirically found that blending in the textureless image, which contains geometric cues from shading, provides implicit guidance to align RGB textures at the correct UV locations (see \cref{sec:ablation}).
As the result, the texture-only optimization modifies the gradient \cref{eq:score_jacobian_chaining} to be
\begin{equation}
    \nabla_\featureimage\loss{SDS}\left(\alpha\featureimage_{\theta_{0}, \beta_{0}, C_{0}, \texture} + \left( 1-\alpha \right) \image_{\theta_{0}, \beta_{0}, C_{0}, \texture=\emptyset}\right),\label{eq:shading}
\end{equation}
where \(\theta_{0}, \beta_{0} \) and \(C_{0}\) denote the initial values of the MLP parameters, the shape parameters and the per-vertex features, respectively, which are held fixed during this step.
Note that control inputs for expression, pose and camera angles (\(\expression\), \(\pose\), \(\azimuth\), and \(\elevation\)) are omitted for brevity.

\paragraph{Geometry-texture dual optimization.} The texture optimization step results in view-consistent images, providing a basis for the geometry optimization,
which aims to adjust the mesh such that the silhouette matches the foreground of the rendered images, thereby eliminating the background in the texture maps.
Specifically, every \(M\) iterations, we render a set of images using the current texture map \(\texture\) from uniformly sampled azimuth angles $\azimuth\in [-30^\circ,30^\circ]$, using the canonical expression and pose parameters \(\expression_{0}\) and \(\pose_{0}\), \ie, \(\image\left( \expression=\expression_{0}, \pose=\pose_{0}, \azimuth, \elevation=0 \right)\).
From these, we precompute a set of reference foreground masks using an off-the-shelf segmentation network~\cite{kirillov2023segany}, denoted as \(S\left( \expression=\expression_{0}, \pose=\pose_{0}, \azimuth, \elevation=0 \right)\), or \(S_{0}\left( a \right)\) for short.
Then we minimize the following loss \wrt \(\shape\), \(\geonet_{\theta}\) and \(C\), in addition to \cref{eq:shading}:
\begin{equation}
\loss{seg}\left( \theta, \shape, C \right) =
\E_{a}\left[\lVert S_{0}\left(a\right) - \texttt{Upsample}(\maskimage_{\theta, \beta, C}\left(\expression=\expression_{0}, \pose=\pose_{0}, a, e=0 \right))\rVert^2\right].
    \label{eq:silhouette_loss}
\end{equation}
Recall that \(\hat{M}_{\theta, \beta, C}\) is the soft silhouette mask from the differentiable rasterizer, which is independent of the texture map \(\texture\).
Also note that the expression and pose parameters are fixed during this step to their canonical values, \ie, \(\expression_{0}\) and \(\pose_{0}\), respectively, to ensure that silhouette change comes from the trainable geometric parameters, \ie \(\theta, \shape,\) and \(C\), only.
The reference foreground masks are updated only once every \(M\) iterations, which is crucial for providing a stable gradient signal despite changing texture.
We alternate between texture-only and dual optimization steps to obtain a converged representation.

\paragraph{Geometry regularization.}
In order to prevent degenerate solutions and ease shape optimization, we apply additional regularization terms to the geometry, constituting $\loss{geom}$.
This regularization is given by:
\begin{align}
   \loss{geom} & = \lambda_1\loss{off} + \lambda_2\loss{lap} + \lambda_3\loss{prior}, \\
   \loss{off} & = \sum\left|\geonet_\theta\left(\left\lbrack T,C\right\rbrack\right) \right|, \\
   \loss{lap} & = \lvert\lap V_{\theta, \shape, C}\rvert,\\
   \loss{prior} & = \lvert \lap V_{\theta, \shape, C} - \lap T \rvert,
\end{align}
where \(\lap\) denotes the Laplacian operator.
Intuitively, $\loss{off}$ regularizes the deviation from the base FLAME template, $\loss{lap}$ encourages smoothness, and $\loss{prior}$ penalizes distortion of geometric details.

\section{Experiments}
\label{sec:experiments}
Below, we show qualitatively and quantitatively that {\ours} can generate high-quality texture and fittingly update the geometry to match diverse text prompts.
For additional results, please see our website at: \texttt{\href{https://www.computationalimaging.org/publications/articulated-diffusion/}{computationalimaging.org/publications/articulated-diffusion/}}

\subsection{Implementation Details}
For all experiments, we use the refined FLAME template mesh provided by~\cite{grassal2022neural}. 
We train for 20,000 iterations, updating segmentation masks every 2,000 iterations after the initial 4,000 iterations of texture-only optimization. Additionally, we decay the $\alpha$ parameter over the course of training. We utilize the same training hyperparameters regardless of input text prompt.
This optimization process takes approximately 2 hours on an NVIDIA RTX6000 GPU. For the full set of hyperparameters,  
please see the supplementary document.

\begin{figure}[t!]
	\includegraphics[width=\textwidth]{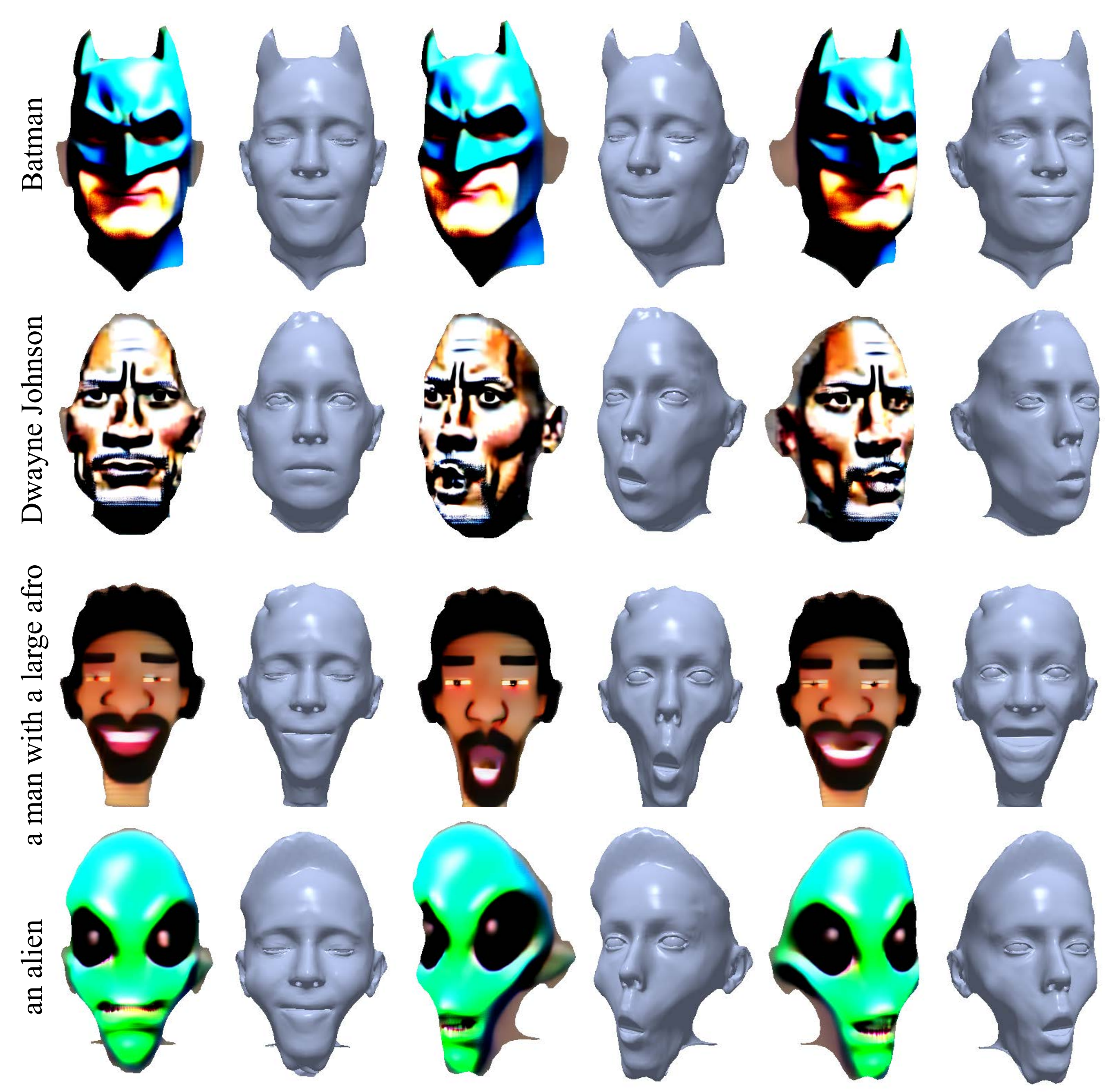}
	\caption{\textbf{Qualitative results.} Our model generates high-quality prompt-consistent texture and geometry. All generated avatars can be efficiently animated using FLAME's linear deformation model. The articulation result of textured and texture-less meshes are highly consistent, thanks to the geometry-aware texture optimization. By using state-of-the-art 2D diffusion models, we can generate a wide variety of text prompts, ranging from humans based on descriptions to fictional figures and well-known celebrities.}
	\label{fig:results}
\end{figure}

\subsection{Articulated head avatars}
\cref{fig:results} shows our renderings and the optimized geometry using different text prompts under varying views and articulations.
Both the texture and geometry show high-level prompt consistency (see the ears in the ``Batman'' example).
Furthermore, the generated avatar faithfully follows the given expression and pose parameters, which can be attributed to our geometry-aware texture optimization.
By utilizing the comprehensive representation power of 2D diffusion models, we are able to generate diverse samples for any description in the latent space.
This includes both general human / humanoid descriptions, and well-known figures.
Please see our website for animated videos driven by varying pose and viewpoint.

\subsection{Comparisons.}
\paragraph{Baselines.}
To the best of our knowledge, apart from two very recent and concurrently developed methods, there is currently no existing work capable of generating articulable head avatars based on a text prompt.
As animation is key to head avatars, we consider two mesh-based baselines that are articulable:
\begin{compactenum}
\item Text2Mesh~\cite{michel2022text2mesh}: similar to our method, it optimizes for both the geometry and appearance. Since it is originally proposed for class-agnostic text-based shape synthesis, we adapt it to head avatar generation by providing the FLAME template as the initial mesh. We modify training to also use the same distribution of camera and FLAME parameters for each sample, resulting in an articulable mesh that can be controlled by the FLAME parameters.
\item ClipFace~\cite{aneja2022clipface}: this concurrent work represents the SOTA result for text-based 3D face generation. It uses a pretrained GAN that generates texture. Given a text prompt, it applies a GAN inversion with Pivotal Tuning Inversion (PTI)~\cite{roich2022pivotal}-like procedure to optimize for the latent code and generator network such that the rendered image matches the given prompt.
\end{compactenum}
Unlike our method, both baselines use the CLIP loss~\cite{radford2021learning} to encourage texture--image alignment.

\begin{figure}[t!]
\centering
	\includegraphics[width=\textwidth]{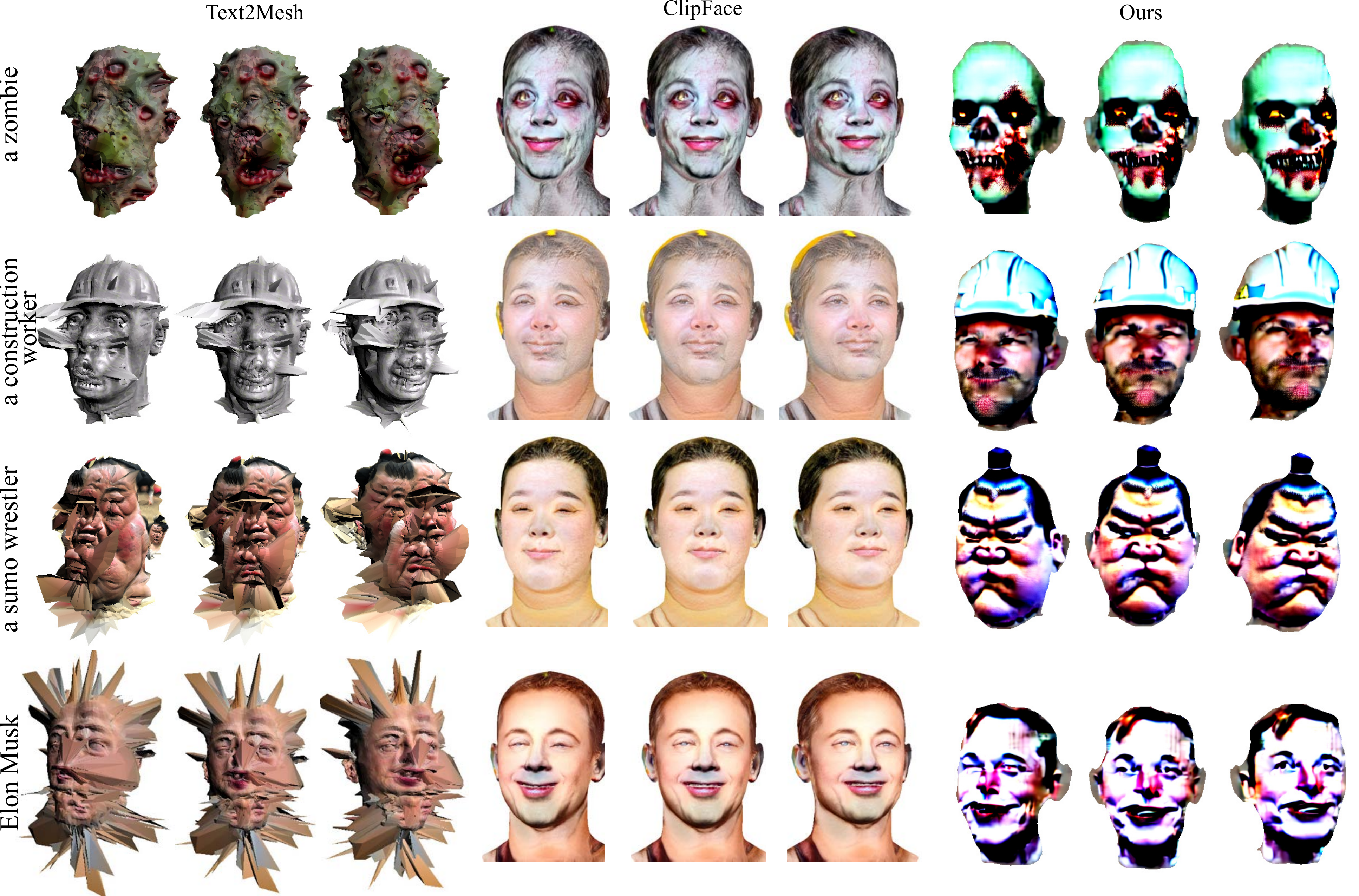}
	\caption{\textbf{Qualitative comparison with existing work.} We show the generated avatars in 3 different views and with an adapted Text2Mesh~\cite{michel2022text2mesh} and a concurrent work ClipFace~\cite{aneja2022clipface}. Both optimize texture based on the FLAME model. While ClipFace fixes the mesh, Text2Mesh also updates the geometry, like \ours{}. Both baselines use the CLIP~\cite{radford2021learning} loss, and demonstrate the opposite ends of spectrum using CLIP directly: with Text2Mesh being under-constrained and yielding non-realistic textures and diverged geometry, whereas ClipFace is dominated by the GAN prior and limited by the representation power of CLIP failing to fit the text prompts accurately. By using SDS with a pretrained diffusion model, our generation is far more diverse than ClipFace.
    }
	\label{fig:comparison}
\end{figure}

\setlength{\intextsep}{0pt}%
\begin{wrapfigure}{R}{0.5\textwidth}
    \centering
    \includegraphics[width=\linewidth]{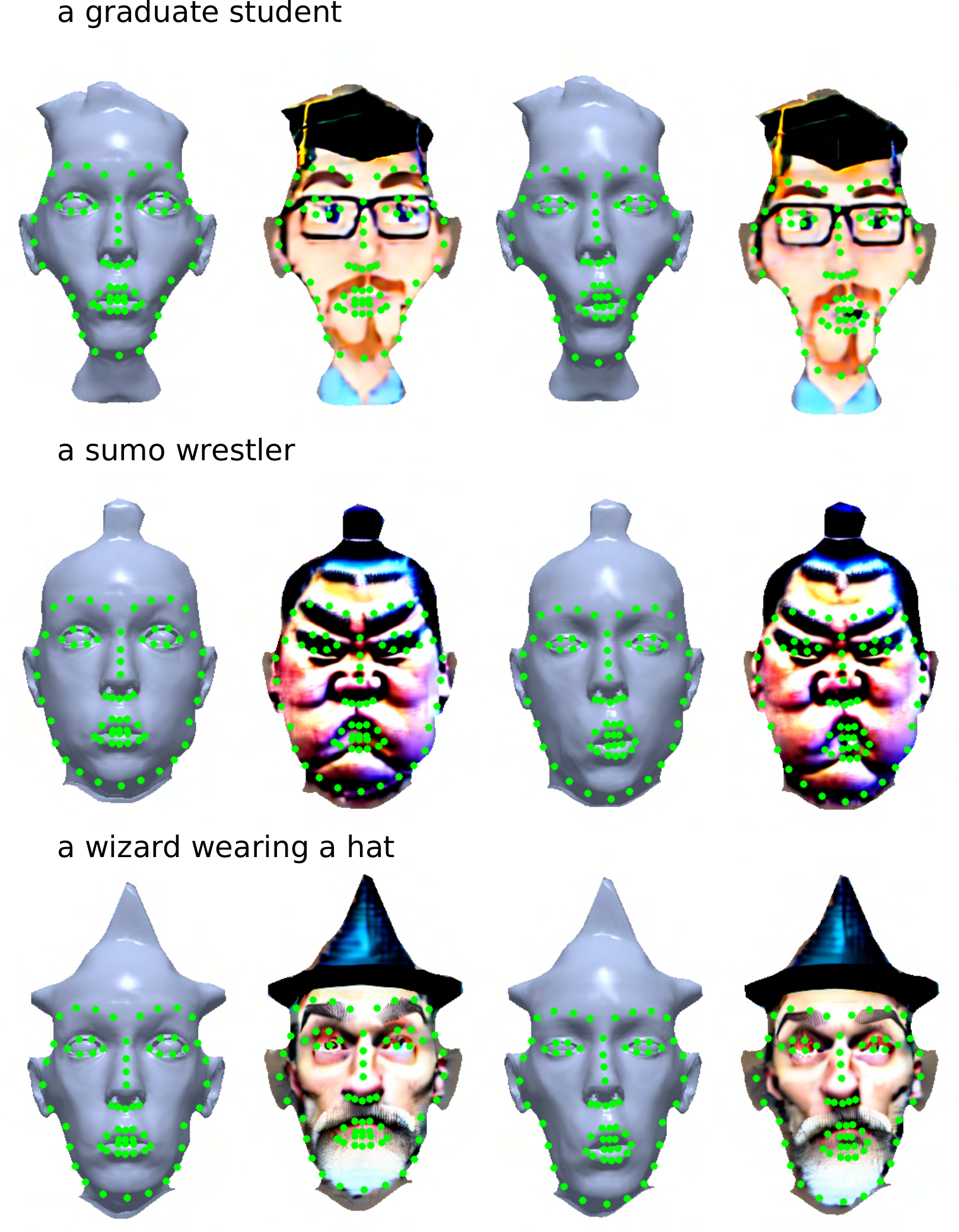}
    \caption{\textbf{Text--geometry and texture--geometry alignment.} With our geometry-consistent dual optimization for texture and shape, we are able to capture characteristic geometry for the given text prompts. Despite the large shape deformation, we maintain good texture-geometry alignment as shown in the visualization of the 3D landmarks, which is critical for realistic animation.}
	\label{fig:geometry}
    \vspace{-1em}
\end{wrapfigure}

\paragraph{Result.}
We summarize the quantitative evaluation and demonstrate the visual quality in \cref{tab:comparison} and \cref{fig:results}.
As shown in \cref{fig:comparison}, Despite having relative good prompt consistency on the rendering (CLIP-R ``Color''), Text2Mesh~\cite{michel2022text2mesh} fails to generate realistic texture for human faces, and the geometry optimization completely diverges.
This is likely because the CLIP loss can not provide reliable signal.
While Text2Mesh introduced various augmention techniques that improved convergence, \eg patch-based jittering, these techniques may have essentially contributed to the multi-face hallucination problem.
ClipFace, operating in a latent space of a GAN pretrained on a filtered FFHQ dataset~\cite{Karras2020stylegan2}, has a strong prior.
As a result, it generates more detailed natural-looking textures.
Additionally, since the GAN training data is pruned and pre-aligned with the FLAME model, the generated results also align well with the 3D facial landmarks.
However, because of this strong prior, it is struggles to handle ``exotic'' prompts or known figures (\eg, ``sumo wrestler'' and ``Elon Musk'').
Furthermore, lacking geometry optimization, ClipFace performs poorly in terms of geometry prompt consistency as shown in the CLIP-R ``Geo'' and in the ``sumo wrestler'' example in \cref{fig:comparison}.

Our method is able to generate high-quality texture and geometry that are consistent with the given text prompt.
As illustrated in \cref{fig:comparison}, our generation covers more diverse prompts, thanks to the use of a diffusion model.
We especially outperform the baselines in geometry--text consistency (see CLIP-R ``Geo'' in \cref{tab:comparison}), as a direct result from the dual geometry--texture optimization.
This advantage is further demonstrated in \cref{fig:geometry}, where we visualize several examples with prompts that require geometry updates.
All of the generated avatars capture the unique geometric features of the given prompt, \eg hat, hair, and face shape, which not only improves the geometry--prompt consistency, but also allows more canvas to paint the texture correspondingly, thereby improving the rendering-prompt consistency as well.
Furthermore, also shown in \cref{fig:geometry}, the projections of the 3D facial landmarks align reasonably well with the 2D landmarks, despite the large changes in the shape, which is a critical property for realistic animation.

\paragraph{Quantitative metrics.}
We evaluate the prompt consistency of the optimized texture and geometry using the CLIP-R Precision metric proposed in DreamFields~\cite{jain2022zero}.
Given a rendered image of an avatar and a collection of candidate prompts, it measures the accuracy with which CLIP retrieves the correct prompt.
Additionally, we also compute this metric on renderings of the textureless mesh to measure the prompt consistency of the geometry.
Similar to ~\cite{poole2022dreamfusion}, we report the result on 3 different CLIP models and their average.
The candidate prompts are created based on 267 random prompts from the DreamFusion gallery page, and an additional 60 prompts used in generating results for our method.
These prompts are included in the supplementary document.
The results are evaluated on 7 prompts and the average is reported.

\begin{table}[htbp]
\small
\centering

\vspace{.8em}
\begin{tabular}{lcccccccc}
	\toprule
	& \multicolumn{2}{c}{CLIP B/32 $\uparrow$} & \multicolumn{2}{c}{CLIP B/16 $\uparrow$} & \multicolumn{2}{c}{CLIP L/14 $\uparrow$} &\multicolumn{2}{c}{Average $\uparrow$}\\
    \cmidrule(lr){2-3} \cmidrule(lr){4-5} \cmidrule(lr){6-7} \cmidrule(lr){8-9}
	& Color & Geo & Color & Geo & Color & Geo & Color & Geo  \\
	\midrule

	Text2Mesh & 74.6 & 8.8 & 73.0 & 0.0 & 66.5 & 0.0 & 77.4 & 2.9 \\
	ClipFace & \textbf{94.5} & 1.7 & 72.4 & 1.0 & 58.2 & 2.7 & 75.0 & 5.4 \\
    \midrule

    Ours (w/o $\loss{seg}$) & 57.1 & 1.7 & 67.7 & 1.0 & 82.8 & 2.7 & 69.2 & 1.8 \\
    Ours (w/o schedule) & 69.0 & \textbf{14.1} & 71.1 & \underline{10.4} & 80.8 & \textbf{9.1} & 73.6 & \textbf{11.2} \\
    Ours (w/o shading) & 74.2 & 11.5 & \textbf{89.2} & 6.8 & \underline{93.7} & 1.0 & \textbf{85.7} & 3.1\\
    \midrule
	Ours & \underline{75.3} & \underline{12.4} & \underline{78.6} & \textbf{12.1} & \textbf{95.8} & \underline{6.8} & \underline{83.2}& \underline{10.4} \\
	\bottomrule
\end{tabular}
\vspace{0.2em}
\caption{\textbf{Quantitative comparison and ablation.} We evaluate the consistency between the avatar renders and the text prompt with CLIP-R using 3 different CLIP models, evaluating both on the textured mesh (see ``Color'' columns) and the textureless mesh (see ``Geo'' columns), which considers the geometry consistency with the text prompt. The best and second best results are highlighted in \textbf{bold} and with \underline{underline}, respectively.}
\label{tab:comparison}
\end{table}

\subsection{Ablation of optimization procedure.}\label{sec:ablation}
One of our main contributions is the dual optimization of geometry and texture as described in \cref{sec:training}.
In this section, we ablate its key components:
\begin{compactenum}
\item Segmentation loss (w/o \(\loss{seg}\)): We omit the geometry update step, which uses \(\loss{seg}\) to supervise the soft silhouette.
\item Scheduled optimization (w/o schedule): We omit the scheduled two-step training that starts with texture-only optimization, and instead apply the dual optimization step from start to finish.
\item Shading cue (w/o shading): We omit the \(\alpha\)-blending described in \cref{eq:shading}.
\end{compactenum}

\begin{figure}[t!]
	\includegraphics[width=\textwidth]{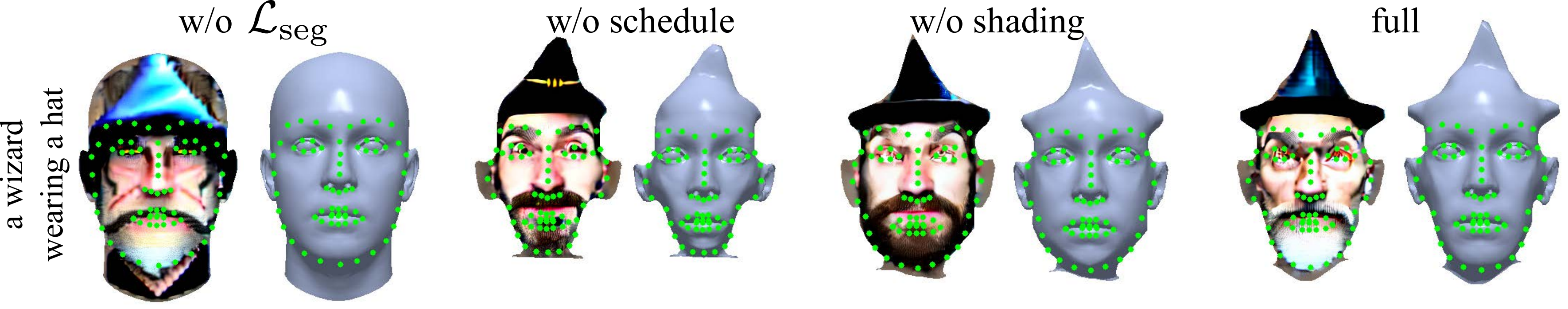}
	\caption{\textbf{Ablation.} We examine the effect of various key components of proposed optimization procedure. Without \(\loss{seg}\) the wizard hat is incorrectly painted on the forehead. Removing scheduled optimization and \(\alpha-\)blending term both lead to worse facial feature alignment (see mouth). The full model aligns the facial feature and matches the prompt closely.
    }
	\label{fig:ablation}
\end{figure}

We report the quantitative results in \cref{tab:comparison} and the visual results in \cref{fig:ablation}.
Without segmentation loss, the geometry remains fixed to the initial FLAME template, yielding much lower CLIP-R scores for the geometry.
The scheduled optimization has a positive effect on the prompt consistency for the texture, as also shown in \cref{fig:ablation}.
We hypothesize that this is because the texture-only training step provides a good initialization for the dual optimization step, thereby improving the stability and convergence of the training.
Finally, removing shading leads to worse geometry--text consistency as well as misaligned facial features as shown in \cref{fig:ablation}, indicating that the shading cue provides useful information for geometry optimization.

\section{Discussion}
\label{sec:discussion}
\paragraph{Limitations and future work.}
Our work is not without limitations.
As is commonly observed in diffusion-based 3D synthesis, the generated images tend to have cartoon-ish stylization and high color saturation, which impacts the realism of the results. Improvement along this axis could potentially enable generation of highly realistic head avatars from text.
Secondly, as the diffusion decoder takes very low-resolution feature maps as input and uses multiple 2D upsampling layers to generate RGB images in individual camera views, the upsampling is not guaranteed to be consistent across different camera views, which can lead to flickering during animation.
In future work, one could consider either fusing rendered images into a new, consistent texture in color space, or applying a color-space diffusion model directly on mesh textures to generate a directly rasterizable result.

\paragraph{Ethical considerations.}
In our proposed method for generating 3D head avatars using a diffusion model and AI, we recognize the importance of ethical considerations and societal impact. While our current results do not pose a significant threat to society, we remain open to discussions and welcome efforts to safeguard against potential risks. It is crucial to acknowledge that the diffusion model, like any AI system, may exhibit biases due to the training data. Therefore, we emphasize the need for ongoing research and development to mitigate biases and ensure fair and responsible use of the technology. By actively addressing these concerns, we aim to contribute to a more comprehensive understanding and responsible deployment of AI-generated 3D avatars.

\paragraph{Conclusion.}
Our work demonstrates a novel approach to generating articulatable 3D head avatars from a variety of text-prompts, which is efficient to render and can be animated in real time. This has the potential to enable applications in virtual reality and personalized digital content creation, and opens up exciting possibilities for immersive experiences in conferencing.
%

%%%%%%%%%%%%%%%%%%%%%%%%%%%%%%%%%%%%%%%%%%%%%%%%%%%%%%%%%%%%

{\small
    \bibliographystyle{abbrvnat}
    \bibliography{references}
}



\section*{Supplementary material}

# Articulated 3D Head Avatar Generation using Text-to-Image Diffusion Models
# –Supplementary Document–

**Alexander W. Bergman**
Stanford University
awb@stanford.edu

**Wang Yifan**
Stanford University
yifan.wang@stanford.edu

**Gordon Wetzstein**
Stanford University
gordonwz@stanford.edu

computationalimaging.org/publications/articulated-diffusion/

## Contents

# 1 Implementation and Evaluation Details

## 1.1 Representation

**Shape.** As described in the main text, the shape is implemented using the human head 3DMM FLAME [7] as a base mesh, and an MLP $\mathcal{G}_\theta$ modeling offsets from this mesh. We use the modified FLAME model proposed in Neural Head Avatars [3]. This modified model uniformly subdivides the faces of the mesh (four-way subdivision), removes the faces belonging to the lower neck region, and adds additional faces close to the mouth cavity. This results in 16227 vertices on the FLAME model mesh rather than 5023. Each vertex has an associated feature vector $c \in \mathbb{R}^d$. We set $d = 32$ for all of our trained models.

The MLP $\mathcal{G}_\theta$ which models offsets takes in a concatenation of all the vertex positions of the neutral FLAME model $T \in \mathbb{R}^{16227 \times 3}$ and their associated vertex features $C \in \mathbb{R}^{16227 \times d}$, and processes each one independently (16227 is the batch dimension), outputting vertex offsets $\in \mathbb{R}^{16227 \times 3}$. This MLP is implemented as a fully connected network with 3 hidden layers, each with hidden activation size 128. It uses $\mathrm{ReLU}$ as the activation function, with the final layer output being put through a $\tanh$ function and scaled by $10^{-1}$. This ensures that the offsets are not too large at initialization so that the mesh that is similar topologically to the FLAME template at initialization.

**Texture.** We model appearance as a feature texture map $\mathcal{T}$ of resolution $512 \times 512 \times 4$, and use the UV coordinates defined with the FLAME model. The features of the texture map are defined in the latent space of Stable Diffusion v1.5 [13], and thus each feature in the texture map is in $\mathbb{R}^4$. Rendered feature images from $\mathcal{T}$ can be decoded into RGB space using the decoder in Stable Diffusion v1.5. This decoder also upsamples the feature images $8x$, resulting in $512 \times 512$ RGB images from a $64 \times 64$ feature image rendering. Note that these feature images can also be transformed into RGB images using the approximate linear transform from the latent space of Stable Diffusion v1.5 to RGB, as defined in [9]. However, we choose to use the provided decoder to upsample feature images in image space.

## 1.2 Optimization

**Initialization.** We initialize the MLP $\mathcal{G}_\theta$ with the default ReLU MLP initialization, and initialize the per-vertex features $C$ in a zero-centered Gaussian distribution with $\sigma = 0.02$. Importantly, in initializing the vertex positions, we add a pre-computed set of offsets to the FLAME mesh in order to enlarge it. This extra offset is computed by applying Laplacian Smoothing [15] with a negative strength, resulting in a smooth mesh with no intersections which is offset roughly along each vertex normal. We include a visualization of the expanded template mesh in Fig. 1. As described in the manuscript, this provides more area for the texture-only optimization to paint texture on, which can then be trimmed by subsequent geometry optimization steps. For example, the enlarged mesh gives room for the texture-only optimization to paint head accessories on, which can then be converted into consistent geometry via the segmentation loss. This is apparent in the examples shown in the main paper where the enlarged scalp results in the ability to paint on head and hat details to be consistent with a specific prompt.

**Score distillation loss.** We use the score distillation loss implementation provided in Score Jacobian Chaining [17]. Unlike DreamFusion [11], which uses the Imagen [14] diffusion model, the Stable Diffusion v1.5 model directly predicts denoised images directly instead of the predicting the noise to subtract from a noised image. In practice, this implementation results in similar performance. Note that this loss function never provides a direct value for text-image similarity, but instead the diffusion model is used to compute gradients with respect to a value which represents this. We do not scale or alter these gradients by any value.

Like in other works using score distillation as a loss function, we find it necessary to use a high classifier-free guidance in order to obtain converged results [11, 17]. We use a classifier-free guidance parameter of 100. As classifier-free guidance is known to produce oversaturated and non-realistic images in generation, we hypothesize that this is the reason for some of the oversaturation and cartoonish effects on some of our generated 3D head avatars.

**Texture only optimization and shading cues.** For texture optimization, we use a learning rate of $8 \times 10^{-3}$. We perform 6,000 iterations of texture-only optimization to start the optimization process, and then 2,000 more iterations after every 4,000 iterations of geometry and texture optimization. In this step, we use only the score distillation loss, and backpropagate from rendered images into the texture $\mathcal{T}$ only. Since gradients propagated into the geometry $(\beta, \theta, C)$ are likely noisy, we manually ensure that appearance gradients are not propagated here.

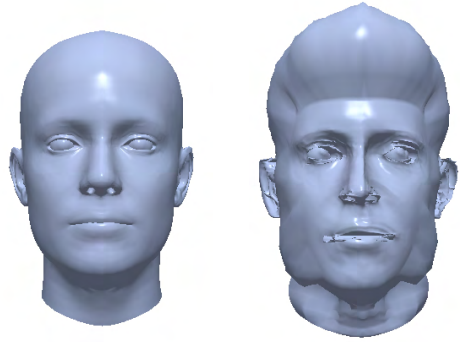

Figure 1: **Enlarged template mesh initialization.** In order to provide a large enough canvas for the score distillation loss to paint a landmark-aligned texture onto, we enlarge the base FLAME template (left) with vertex offsets at initialization (right).

As described in the main text, to encourage semantic features to be generated in the correct locations (i.e. mouth on the FLAME model mouth), the rendered textured images are blended with textureless renders of the offset mesh. These are obtained as in Eq. 5, and the amount of influence from the textureless image is controlled by $\alpha$. We tune this parameter over the course of training as follows: for the first 4,000 iterations, $\alpha = \frac{\texttt{curriter}}{4000}$, i.e. linearly increasing to 1. After this, we set $\alpha$ to be uniformly distributed between 0.6 and 1.0, leading to more contribution from the rendered feature image on average. This schedule encourages the texture to align features with the correct landmarks early in optimization, so that the score distillation gets caught in a local minimum with features aligned. Then, later in optimization, lower $\alpha$ allows the texture to focus on more detailed appearance features without accounting for the contribution from the textureless image, which is not used in rendering final images or videos.

**Geometry optimization.** Following the texture-only optimization step, we perform a joint geometry and texture optimization step. In this case, we use a learning rate of $8 \times 10^{-3}$ for texture parameters $\mathcal{T}$, and a learning rate of $1 \times 10^{-4}$ for geometry paramters $\beta, \theta, C$. We propagate gradients from the score distillation loss only into $\mathcal{T}$, as in the texture-only optimization step. However, we propagate gradients from the segmentation loss $\mathcal{L}_\text{seg}$ and regularization loss $\mathcal{L}_\text{geom}$ into the parameters modeling geometry: $\beta, \theta, C$.

After every 2,000 iterations of geometry optimization, we compute a look-up table of approximate segmentation masks for views which can be observed during training. For this, we set the expression and pose parameters to be in the neutral position, and vary the viewpoint direction, computing a separate segmentation mask for each elevation and azimuth $e, a$ which can be sampled at training with $1°$ granularity. In our results, we do not vary $e$, and sample $a \in [-30°, 30°]$, resulting in us computing 61 segmentation masks for each $1°$ interval in $[-30°, 30°]$. We compute these segmentation masks using the Segment Anything [5] model. This model returns various segmentation masks at different levels of specificity (i.e., it can segment only hat or hair). We take the segmentation mask which has the largest amount of pixels, and includes pixels known to intersect with the center of the FLAME template. This way, we are able to obtain a segmentation mask of the entire head, including accessories, and use it as supervision in $\mathcal{L}_\text{seg}$.

We compute $\mathcal{L}_\text{seg}$ using the stored segmentation masks (updated every 2,000 iterations) and upsampled, rasterized soft masks at each iteration, as described in Eq. 6 of the main paper. We multiply this loss by a factor $\lambda = 1000$.

**Geometry regularization.** We apply $\mathcal{L}_\text{geom}$ as described in Eq. 7. We use the following parameters for weighting the relative regularization terms: $\lambda_1 = 0.5, \lambda_2 = 5000, \lambda_3 = 5000$. We implement $\mathcal{L}_\text{off}$ by computing offsets from the base FLAME mesh at a specific expression, rather than the expanded FLAME mesh used at initialization. We compute $\mathcal{L}_\text{lap}$ using PyTorch3D [12] mesh Laplacian smoothing, where the Laplacian is computed with uniform weights. Finally, we compute $\mathcal{L}_\text{prior}$ by computing a weighted L2 difference with the mesh Laplacian of the ground truth FLAME model mesh (without any enlargement). We weight the parts of the mesh in computing this Laplacian difference in the following way: all regions are weighted equally with weight 1, but the "scalp" region is weighted $10^{-1}$, the "face" region is weighted $5 \times 10^{-1}$, and the "forehead" region is weighted

15

$5 \times 10^{-1}$. This is because these regions have the most area for accessories or complex geometry on the head, and thus should not align as closely with the FLAME template.

**Viewpoint and pose sampling.** As we focus on articulation of the facial expression, we sample viewpoints facing the front of the FLAME model with 0 elevation and azimuth in $[-30°, 30°]$. All camera positions are taken at a radius of 0.7 from the origin, where the FLAME template is centered. This allows the views to focus on the front of the face, and not worry about using text prompt conditioning to accurately fit the back of the head and prevent multiple faces from appearing [11]. We randomly vary poses during training. We sample these poses from a dataset obtained by running the tracking on the first sample video provided in the NHA [3] repository. This corresponds to 1496 different expressions used during training. We use the tracking from the second provided as video as validation expression parameters to drive our generated models.

During training, in order to improve the quality of the generated results, we pre-pend "a photo of the face of" to each of the generated prompts.

**Rasterization.** As described in the text, we use a differentiable mesh rasterizer based on SoftRasterizer [8] in Pytorch3D [12] to rasterize images and text. For rasterizing feature images to be passed into the score distillation loss, we perform a "hard" rasterization which does not blur mesh faces and softly blend overlapping rasterized feature values. This results in gradients only being passed into $\mathcal{T}$, and a hard silhouette mask which does not pass gradients into $\beta, \theta, C$. We also use this setting to rasterize the textureless shading guidance mesh image. When rasterizing soft masks to be used in geometry optimization, we use rasterizer parameters $\sigma = 10^{-4}$, $\gamma = 10^{-4}$, and 75 faces per pixel. This ensures that the rasterization of the mask is differentiable with respect to the vertex positions. Note that when optimizing both texture and geometry, we must perform two rasterizations - one with hard blending and one with the differentiable soft blending.

Since the rasterization of features is performed at the low resolution of $64 \times 64$, there can be faces which are never observed in training pose variation, resulting in uncanny visualization when the heads are animated into new expressions after optimization. Thus, we rasterize feature images at a resolution of $512 \times 512$, and bilinearly downsample them into $64 \times 64$ resolution before passing them into the decoder to be again upsampled. This downsampling and blurring in rendered features encourages gradients to be passed into all features located in the texture, as the higher resolution rasterization does not suffer from aliasing. Since the blurring is done in feature-space, it does not result in as blurry of a resulting image after being decoded. Additionally, to prevent the model from painting the background into $\mathcal{T}$, we vary the background at each training iteration between a randomly sampled uniform color and a checkerboard with two randomly sampled colors and boxes of a randomly sampled size between 15 and 25 pixels.

**Computational Complexity.** We train each model using an NVIDIA RTX6000 GPU. Each model consumes approximately 12GB of VRAM and trains for approximately 1.5 hours, corresponding to 20,000 iterations. We used a total of 4 RTX6000 GPUs in training models and preparing results for the paper.

## 1.3 Rendering videos.

When rendering videos, we are able to drive the optimized avatar representation using new FLAME expression and pose parameters. We retain the geometry by keeping the parameters $\beta, \theta, C$ which were optimized for during training. As the decoder performs an $8x$ up-sampling operation, converting $64 \times 64$ feature images to $512 \times 512$ RGB images, unwanted flickering artifacts can be visible as small changes in the feature value are propagated into a larger region of the image. Thus, we rasterize features at $512 \times 512$ resolution, and bi-linearly downsample these feature images to $64 \times 64$ before decoding them into $512 \times 512$ images.

## 1.4 Evaluation

We evaluate using the CLIP-R Precision metric proposed in Dream Fields [4]. We similarly evaluate this metric using three different CLIP models, on both rendered RGB images and rendered textureless offset mesh images. We compute the recall metric using 100 rendered images for each target prompt. Each of these images are rendered with randomly selected FLAME expression and pose parameters

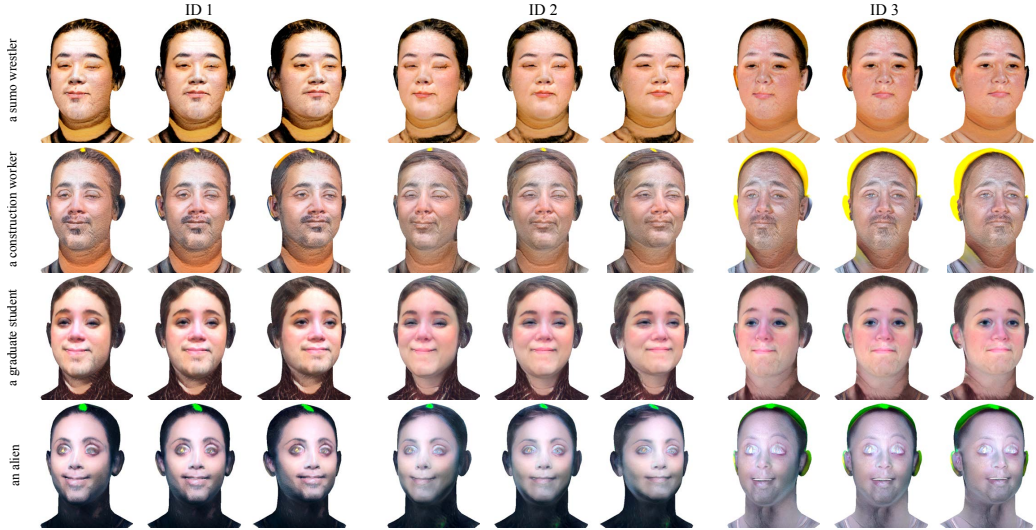

Figure 2: **Additional ClipFace results.** We show results from ClipFace for prompts shown in the main paper with different initial latent codes in the validation set. Additionally, we show two different prompts: "a graduate student" and "an alien". The different latent codes change the generated identity slightly for each given prompt.

from the training dataset, and randomly selected views where azimuth $a \in [-30°, 30°]$. We add random noise as the background (instead of having a uniform color) to prevent the CLIP model from focusing exclusively on the silhouette of the head and biasing the results to head-related prompts. We do the same for the rendered textureless images of the mesh in order to measure geometric consistency with the prompts.

We evaluate CLIP-R Precision for 7 prompts. These prompts are: "an alien", "Elon Musk", "a zombie", "a sumo wrestler", "a construction worker", "a graduate student", "a man with a large afro". We pre-pend the modifier "a photo of the face of" to each of these prompts. As potential false positives, we include 267 randomly scraped prompts from the DreamFusion gallery. However, these prompts come from a very different distribution than the list of head prompts which we use, and thus CLIP obtains almost perfect recall by just focusing on the fact that each image is a head. Thus, we add an additional 60 prompts, many of which were used to generate results for our method. Each of these prompts is also pre-pended with "a photo of the face of." The list of prompts is here: "a man with a large afro", "a man wearing a wearing a top hat", "a man wearing a hat", "a wizard wearing a hat", "a man with large spiky hair", "a zombie", "an alien", "Batman", "Keanu Reeves", "Rihanna", "Shaq", "Jackie Chan", "a senior citizen", "a doctor", "a graduate student", "a sumo wrestler", "a construction worker", "Leonardo DiCaprio", "Cristiano Ronaldo", "Ryan Gosling", "Margot Robbie", "Elon Musk", "Dwayne Johnson", "Shrek", "a lawyer", "a teenager", "a gardener", "a pilot", "a firefighter", "a soldier", "a scientist", "a police officer", "a Jedi", "Godzilla", "James Bond", "Bill Gates", "Lionel Messi", "LeBron James", "Roger Federer", "an astronaut", "a man with a large chin", "a man wearing a baseball hat", "a female scientist", "Santa Claus", "Hulk", "an anime character", "Cloud Strife", "a man with a swollen face", "Jeff Bezos", "Mark Zuckerberg".

**Baselines.** As described in the main text, we compare to the baselines of Text2Mesh [10] and ClipFace [2]. Both of these are capable of converting text prompts into 3D representations (and with some modifications, head avatars). These methods both use CLIP. Details of their implementation and evaluation is included in the following two paragraphs.

**Text2Mesh.** We use the official Text2Mesh repository, and make a few edits to ensure that the representation is able to create a 3D Head Avatar comparable to ours. Specifically, we use the same FLAME model from NHA and our method (with the expanded number of vertices) as the base mesh on which Text2Mesh operates. Additionally, we modify the training scheme to sample viewpoint and FLAME expression and pose parameters to exactly match the training of our method. We keep other training details, such as the loss formulation with CLIP and rendered image augmentations, the same as the default Text2Mesh.

17

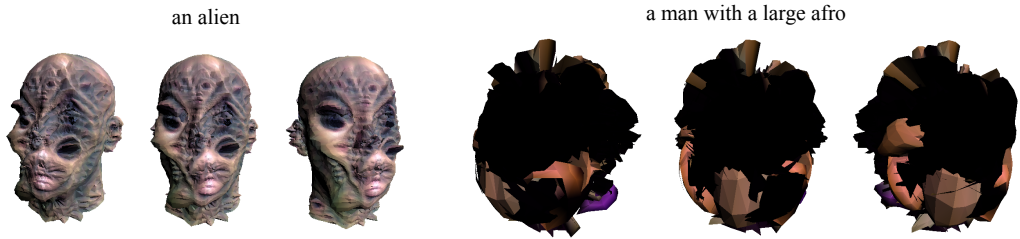

Figure 3: **Additional Text2Mesh results.** We show additional results from Text2Mesh on different evaluation prompts. As with the prompts shown in the main paper, Text2Mesh tends to produce results which appear to contain textures resembling the prompt but are not globally cohesive, such as in the "alien" example. For some prompts, such as "a man with a large afro", the lack of geometric regularization results in divergence.

**ClipFace.** We obtained an implementation of ClipFace from the authors along with a pre-trained GAN model, which we used directly for prompt-specific optimization and evaluation. We additionally obtained training and validation latent codes (each representing a different identity in the ClipFace GAN) from the authors for training the mapping network for a specific prompt. This mapping network was then trained for each specific prompt in the 7 previously mentioned prompts we use for evaluation. We train each of the mapping networks using the default hyperparameters (and FLAME model) for 25,704 iterations, at which point the validation samples appeared to resemble the specific text prompt. For rendering images, we select a random latent code (representing a random latent identity) in the validation set. This identity is then transformed to be compatible with the text prompt by passing it through the trained mapping network for that prompt. We use default values for evaluating this network, and sample the initialization latent codes from the provided validation dataset by the authors.

## 2 Additional Results and Discussion

For additional results of our method, such as rendered videos demonstrating articulation and multi-view rendering, please see our website at:

```
computationalimaging.org/publications/articulated-diffusion/
```

**ClipFace.** In Fig. 2, we include additional results from ClipFace. These results show more examples for various evaluation prompts with different latent vectors used as initial input, corresponding to different person identities in the ClipFace GAN. We can see that using different initial latent codes makes minor differences in the output generated result, but the results remain consistent with the prompt. However, all of the generated results still suffer from the same types of artifacts in ClipFace - lack of flexibility in the prior to model diverse, humanoid subjects, and lack of geometry optimization, as the result is constrained to generating the texture on the FLAME mesh. We note that ClipFace provides other functionality, such as text-driven expression generation. This is an orthogonal contribution to ours, as we use the text to only generate appearance and shape for an avatar for which expression can be controlled.

**Text2Mesh.** We include additional Text2Mesh results in Fig. 3. Similar to the results in the main paper, Text2Mesh produces textures resembling the prompt but are not globally cohesive. This is because in order to get CLIP supervision on 3D representations to converge, methods like Text2Mesh require rendering multiple cropped and augmented images at every training iteration. This is well studied as a method used in order to avoid local minima in the CLIP latent space [6]. Additionally, the lack of geometric regularization results in unstable training for some prompts, as is visible in Fig. 3. The textures resemble the final prompt, but do not generate cohesive nor high-quality results.

**Score Jacobian Chaining.** We emphasize that while methods like DreamFusion [11] and Score Jacobian Chaining [17] are able to convert text prompts into 3D representations, they are **not** compatible with 3D head avatar generation. This is because these methods fundamentally have no control over what is generated besides the prompt. For example, in Fig. 4, we generate results for the prompts "a man wearing a tophat" and "Drake" using the same prompt pre-conditioning, i.e. "a photo

18

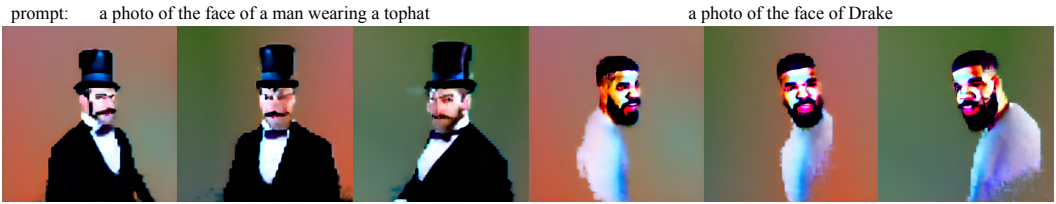

Figure 4: **Score Jacobian Chaining results.** We show results generated by generic text-to-3D methods such as Score Jacobian Chaining, which also uses the Stable Diffusion v1.5 model. These results generated are not controllable or compatible with a 3DMM like FLAME. We note that despite text conditioning the result to generate a face, the method still generates a full upper body. Our method generates results which are qualitatively similar in quality, but are guaranteed to be articulable 3D head avatars.

of the face of". While these methods are able to produce a 3D representation, this representation is not aligned with a 3DMM such as FLAME and doing so is not a trivial task. We emphasize that our method produces 3D results which are qualitatively on-par with these methods, but retain the ability to control what is generated: an articulable 3D head avatar.

**Different diffusion models.** We emphasize that our formulation is not dependent on the specific diffusion model used, i.e. Stable Diffusion v1.5. Other diffusion models can be used in the score distillation loss formulation, often producing better results. For example, the recently released DeepFloyd-IF [1] has led to significant improvements in open source, non-articulated text-to-3D projects [16], including significantly reducing the saturation of generated results. This is a fruitful direction of future work in generating higher quality 3D head avatars from text.



\end{document}